\documentclass[11pt]{article}

\usepackage{titlesec}
\titleformat{\chapter}{}{}{0em}{\bf\LARGE}

\usepackage{geometry}
\usepackage{blindtext}
\usepackage{multicol}
\usepackage{algorithm}
\usepackage{algpseudocode}
\usepackage{amsmath}
\usepackage{graphicx}
\usepackage{authblk}
\usepackage{fancyhdr}
\usepackage{lineno,hyperref}
\usepackage{tikz}
\usepackage{array}

\newcommand\blfootnote[1]{%
  \begingroup
  \renewcommand\thefootnote{}\footnote{#1}%
  \addtocounter{footnote}{-1}%
  \endgroup
}

\providecommand{\keywords}[1]
{
  \small	
  \textbf{\textit{Keywords---}} #1
}
\date{}
\title{Newton methods based convolution neural networks using parallel processing}

\begin{document}

\author{Ujjwal Thakur$^1$ , Anuj Sharma$^1$ \\ Department of Computer Science and Applications,
Panjab University, Chandigarh, India  }

\maketitle

\begin{centering}{\section*{Abstract} } \end{centering}

{The Training of convolutional neural networks is a high-dimensional and non-convex optimization problem. At present, it is inefficient in situations where parametric learning rates can not be confidently set. Some past works have introduced Newton's methods for training deep neural networks. Newton's methods for convolutional neural networks involve complicated operations. Finding the Hessian matrix in second-order methods becomes very complex as we mainly use the finite differences method with the image data. Newton's methods for convolutional neural networks deal with this by using the sub-sampled Hessian Newton methods. In this paper, we have used the complete data instead of the sub-sampled methods that only handle partial data at a time. Further, we have used parallel processing instead of serial processing in mini-batch computations. The results obtained using parallel processing in this study, outperform the time taken by the previous approach.  
}

\keywords{Deep Learning, CNN, Parallel Processing, Image classification, Neural Networks }

\blfootnote{
 Ujjwal Thakur
\\  E-mail: ujjwal4assign@gmail.com
 
}
\blfootnote{
  Anuj Sharma \\
E-mail: anujs@pu.ac.in \\
Homepage: https://anuj-sharma.in
}

\section{Introduction}
{

  We humans have set a goal of creating or more precisely designing non-biological intelligence for a long time, this goal can be traced back to centuries, under some definitions, even by millennia. That is long before a formal establishment of the field of artificial intelligence at a workshop in Dartmouth College in 1956 \cite{termAI}. There are multiple definitions of AI, Bellman in 1978 defined AI as ``The automation of activities we associate with human thinking, activities such as decision-making, problem-solving, learning..'' \cite{Bellman}, Winston in 1984 stated that AI is 'the study of ideas that enable computers to be intelligent'\cite{Winston}, according to  Charniak and McDermott in  1985 AI is “the study of mental faculties through the use of computational methods”\cite{Charniak} while Dean et al. in 1995  defined AI as “the design and study of computer programs that behave intelligently” \cite{Dean}. Definitions might differ but the central idea of AI is that we want to design a non-biological machine that can mimic the cognitive intelligence we have in some biological beings. In other terms, it can also be said that any machine that exhibits traits such as learning and problem-solving is an AI. One thing is clear, the developments in the field of artificial intelligence will dramatically influence human life, and society and will change the world.
In past, traditional non-AI algorithms are bad at making environment-dependent decisions. Traditional algorithms can check for millions of predefined cases but when even a single undefined case arises they fail. Due to this limitation, these algorithms can't be used in problems like driving cars, natural language recognition, detecting fake news and many more problems. That is why we need AI algorithms, which can adapt according to environmental needs. \par
Till now, we have discussed about making machines able to mimic human intelligence. A technique used for this purpose is called machine learning. Machine learning is an application of AI that lets machines learn on the go and improve with the experience. Machine learning algorithms are designed in such a way that they can collect data while running and use that data to improve themselves in solving upcoming problems. Further going into machine learning algorithms,they can be categorize as supervised machine learning algorithms, unsupervised machine learning algorithms, semi-supervised machine learning algorithms, and reinforcement machine learning algorithms.
Supervised machine learning algorithms analyze labeled training data which is later used for mapping new tasks. While in unsupervised machine learning algorithms data is unlabeled and the goal is to learn more about the underlying data. Semi-supervised machine learning algorithms have small labeled training data and large unlabeled data. Reinforcement machine learning algorithms interact with the environment by doing some action and get errors or success. These results are then used to maximize the performance of an upcoming task. In machine learning, there are numerous classes of models out of which neural networks are a specific set of algorithms that have transformed Machine Learning. \par
A biological brain process information using networks of neurons. These neurons receive input, process it, and give output to the connecting neurons in form of an electrical signal. We have mimicked this concept of neurons in a neural network.
Neural networks are a set of machine learning algorithms that are modeled on the biological brain. In these algorithms, artificial neurons are the core computational unit that collects and classifies incoming data. NNs have multi-layer networks of neurons. A neuron acts like a function that computes the weighted average of the incoming data and passes the information. Neurons collectively used are called a layer. When many different kinds of layers are stacked together this is called a deep neural network.

A deep neural network is made up of an input layer one or more hidden layers and an output layer. As the name suggests, input is passed through the input layer which is generally a multidimensional vector of numbers. Then these inputs are fed into the hidden layer. The hidden layer then applies some predecided operations to the incoming data and sends data to the next layer which might be another hidden layer or put layer. The hidden layer makes changes to data in such a way that the final output is as optimized as possible. This process of multiple hidden layers used together to learn more about input data is called deep learning.

Convolutional Neural Networks are feed-forward networks. These are specific types of deep neural networks that are designed explicitly keeping the static image in mind. This allows for a few tweaks to get more efficient than general NN when working with images. The most concerning weakness of standard NNs is that they struggle with the computational complexity required to process image data. It's not a hard and fast rule though, for smaller data most standard NN can work.

As we discussed in NNs layers are stacked together, CNN uses mainly three types of layers to build convolutional network architecture, convolutional layer, pooling layer, and fully-connected layer. We know that training of convolutional network is non-convex optimization problem\cite{nonconvex,distcj}. Convolution Neural Networks (CNN) have shown great potential in the field of image processing\cite{inimage}. As deep learning involves solving a difficult non-convex optimization problem, Stochastic Gradient (SG) methods and their different variations are frequently used for solving these problems\cite{Alex2012,simonyan2015deep}. While SG is most suitable, it may not be efficient in certain conditions. Due to this reason training of deep networks using second-order methods has got plenty of interest in the past\cite{Martens2010}. In the literature newton method for training neural networks has been studied (e.g., \cite{pmlr-v70-botev17a,large,10.5555/3104322.3104416}). Line search Newton-CG also called truncated Newton Method is suitable to achieve second-order method on high dimensional optimization has been studied for decades by Nocedal and Wright\cite{NoceWrig06}. The newton-CG method does not require explicit knowledge of Hessian matrix\cite{He2017DistributedHO}, it requires only the Hessian-vector product for any given vector\cite{exple}. Hessian-vector products have a use case in training a deep neural network, also known as Hessian-free optimization. Recently, Newton's methods have been investigated as an alternative optimization technique, but nearly all existing studies consider only fully-connected feed-forward neural networks\cite{cjlin}. Newton's methods for CNN involve complicated operations due to this limited researchers have conducted a thorough investigation. One of the major works in this direction is the introduction of Newton methods in CNN for optimization\cite{cjlin}. There are many reasons to work further in this direction. At first, it is generally more substantial to apply weight updates derived from second-order methods in terms of optimization aspects. Meanwhile, it takes roughly the same time to obtain curvature-vector products\cite{kiros} and compute the gradient which makes it possible to use the second-order method on large scale model\cite{large}. This second-order method has shown better results than traditional stochastic gradient (SG) methods \cite{cjlin}. But it needs to find the Hessian matrix. Finding the Hessian matrix becomes very complex as we might resort to the finite differences method while dealing with the Image data \cite{cjlin}. Another issue is that if we have n dimensions, the Hessian will need \(O(n^2)\) space and computational complexity for n $\times$ n matrix. \cite{cjlin} CJ Lin et. al deals with this by using the Hessian-Free Newton Methods technique introduced by \cite{Martens2010}. Hessian-Free Newton's methods are a little misleading in a way that it tricks us into believing that we will never have to find the Hessian matrix. But that's not the case, We will need to find it once for every step we take. Sub-sampled Hessian Newton methods that have been proposed to save some more space and computation power \cite{7226530}. This trade the next step's direction accuracy for less computational needs. In this work, we took the whole data while making the process more efficient by applying process parallelism. This will give us a Gauss-Newton matrix that is not from sub-sampled data and our Hessian matrix will be more accurate in predicting the next step direction for minimizing. Sub-sampled Hessian Newton methods were used because of the high computational cost in the first place, that is why during implementation, we further divided the data into mini-batches. These mini-batches are then used for getting our matrix. As in mini-batch processing, each batch is independent of other batches, so serial processing implementation is inefficient as the whole computation power of processors is not used. If batches are processed in parallel, tasks will get completed faster. To apply this process parallelism we used programming language tools like threading and multiprocessing. Using multiprocessing\cite{multiprocessing} library we used multiple cores simultaneously.
\subsection{This paper is organized as follows} In Section 1, We introduce CNN and ongoing work in the field. In Section 2, We list our contributions and algorithm. In Section 3, We have presented our experimental results and comparisons. Section 4 concludes this work and discusses about future work. }

\section{Literature Of CNN }

Hubel and Wiesel in 1962 and 1965  \cite{Hubel1962} \cite{Hubel1965}
proposed a model of the visual nervous system, according to that model, the neural network in the visual cortex has a hierarchical structure as LGB (lateral geniculate body), simple cells, complex cells, lower order hyper-complex cells, higher order hyper-complex cells respectively in the hierarchy. In this hierarchy, a cell in a higher stage generally tends to respond selectively to a more complected feature of the stimulus pattern and at the same time, has a larger receptive field and is more insensitive to the shift in the position of the stimulus pattern. \par
Later on, this model couldn't hold its original form but if we only consider the main stream of information flow in the visual system, this model doesn't contradict later work in this field. Findings from this paper helped in making a non-biological neural network that doesn't get affected by the size and position of the pattern.\par
Several models with a target of the ability to recognize patterns like biological minds were purposed. Matthew Kaberisky in 1960 in his book `` A proposed model for visual Information processing in the human Brain'' \cite{Matthew} gave such a model. Similarly, Frank Rosenblatt 1962 in his book'' ``Principles of Neurodynamics: Perceptions and theory of brain mechanisms'' \cite{Frank} also gave a model for pattern recognition. Another model by Giebel in 1971 in his book ``Pattern recognition in biological and technical systems'' \cite{Giebel} also tried to achieve a similar goal.\par
However, the model on the mechanism of feature extraction in the visual nervous system purposed by the author K. Fukushima in 1970 and 1971 \cite{Fukushima1970} \cite{Fukushima1971} would be one of the examples which showed the capabilities of multilayered neural network. But, in this model, the synaptic connections between neurons were fixed and plastic modifications of the synapses were not considered. This paper was succeeded by K. Fukushima in 1975 with the paper ``Cognitron : A self-organized multi-layer neural network''  \cite{Fukushima1975CognitronAS}. In this paper new hypothesis for the organization of synapses between neurons was proposed. By proposing this hypothesis a new multilayered neural network that is effectively organized was introduced that was called ``Cognitron''. \par
However, the responses of all these models were severely affected by the shift in position and/or introduction of distortion in the shape of the input pattern. This was overcome by the neural network model ``Neocognitron'' proposed by K. Fukushima in 1980 in his paper ``Neocognitron : A self-organizing neural network model for a mechanism of pattern recognition unaffected by shift in position'' \cite{FUKUSHIMA1982455}. This paper proposed an improved neural network model that after completion of self-organization, has a structure similar to the hierarchy model of the visual nervous system of the vertebrate. The neural network proposed in the paper was self-organized by using unsupervised learning and acquired the ability to recognize stimulus patterns based on the geometrical similarities of their shapes without getting affected by their position or by small distortion of their shapes.\par
Backpropagation is one of the core components of CNN. Various derivations have been reported in different contexts by Parker in 1985 \cite{parker1985learning} and  Werbos in 1974 \cite{Werbos}. The simplest one is given by  Rumelhart et. al. in 1986 \cite{Rumelhart}. Le Cun in 1986  used local criteria attached to each unit which are minimized locally \cite{LeCun1986}. Several variation of this algorithm were given by le cun in 1985 and 1986 \cite{LeCun1985} \cite{LeCun1986}. Le Cun in  1988 proposed a derivation of back-propagation based on the Lagrangian formalism \cite{LeCun1988}. This formalism in back-propagation is inspired by optimal control. Variational calculus is a continuous version of optimal control that uses the method of Lagrange multiplier to find the optimal values of a set of control variables. Variational calculus and it's extensions are in-fact the basis of most work in optimal control [Noton in 1965\cite{noton1965}; Athans and Falb in 1966 \cite{athans1966}; Bryson and Ho in 1969 \cite{bryson1969}]. Algorithms given in this formalism resemble back-propagation. \par

The central problem that back-propagation solves is the evaluation of the influence of a parameter on a function whose computation involves several elementary steps. The solution to this problem was given by chain rule, but back-propagation exploits the particular form of the function used at each step to provide an elegant and local procedure. Following the variational formalism of Lagrange, Pontryagin has shown in the late 1950s how to formulate this problem using a single energy-like Hamiltonian function. An extensive treatment of Pontryagin's minimum principle can be found in  Athanas's and  Falin's work in 1966 \cite{athans1966}. For the problem of simple feed-forward multilayered networks, the full generality of Pontryagin's result, even of variational calculus, is not needed. Only the standard Lagrange's multiplier method will be used. Some of the applications and algorithms described in the optimal control literate so closely resemble back-propagation that one could credit Pontryagin for its discovery. Although their description was in the framework of optimal control, not machine learning, the resulting procedure is identical to back-propagation. Bryson and Ho in 1969 \cite{bryson1969} was the first to give a description of back-propagation as we know it, Although the idea of back-propagating derivatives is much older especially for continuous line systems \cite{athans1966}. The idea of connecting units to local receptive fields goes to the early 60s. Local connections have been used many times in neural models of visual learning \cite{Fukushima1975CognitronAS} \cite{LeCun1986} \cite{LeCun1989}, with local receptive fields\cite{Mozer1991}, neurons can learn to extract elementary visual features as oriented edges, end-points, corners. These features are then combined by the subsequent layers in order to detect higher-order features. Distortion or shifts in the input can cause the position of salient features to vary. In addition, elementary feature detectors that are useful on one of the images are likely to be useful across the entire image. This knowledge can be applied by forcing a set of units, whose receptive fields are located at different places on the image, to have identical weight vectors \cite{FUKUSHIMA1982455} \cite{Rumelhart1986} \cite{LeCun1989}. \par
The convolutional sub-sampling combination, inspired by Hubel and Wiesel mentions of simple and  complex cells was implementer in Fukuyama's Reconnoitering, though no globally supervised learning procedure such as back-propagation was available then such as back-propagation.\par
Since all the weights are learned with back-propagation, convolutional networks can be seen as synthesizing their own feature extractor and tuning them to the task at hand. The weight sharing technique has the interesting side effect of reducing the number of free parameters., thereby reducing the ``capacity '' of the machine and reducing the gap between test error and training error \cite{LeCun1989}. Fixed-size CNN have been applied to many applications, handwriting recognition \cite{Lecun1990} \cite{Martin1993}, machine-printed character recognition \cite{Wang1993}, as well as online hand-writing recognition \cite{Bengio1995}.

An earlier version of this formalism was presented by  Fogelman-Soulie et al in 1986 \cite{Fogelman-Soulie1986}. The credit for establishing CNN as a preferred method for pattern recognition in static images can be given to  `` Object Recognition with gradient-based learning''  \cite{LeCun1999}\cite{LeCun1989}. This used learning about connecting to local receptive fields on input from Hubel and Wiesel \cite{Hubel1962}. Findings from Fukushima \cite{Fukushima1975CognitronAS}), Lecun \cite{LeCun1986}, Fukushima and Miyake \cite{FUKUSHIMA1982455}, Rumelhart,Hinton and Williams \cite{Rumelhart1986}.

K. Chellapilla et al. in 2006 did the first GPU implementation of CNN in the paper `` High Performance Convolutional Neural Networks for Document Processing''\cite{Chellapilla2006}. This implementation was 4 times faster than multi-layer perceptron implementation. But this implementation and some later ones by Uetz and Behnke \cite{Utez2009} and Strigl et al. \cite{Strigl2010}  were hard-coded for specific GPU hardware constraints or used general purpose libraries. Then Dan C. Ciresan, Ueli Meier, Jonathan Masci, Luca M. Gambardella and  Jurgen Schmidhuber in their paper ``Flexible, High-Performance Convolutional Neural Networks for Image Classification'' presented CNN where a flexible and fully online was implemented \cite{Ciresan2011}. This implementation of CNN on GPUs was 10 to 60 times faster than a compiler-optimized CPU version. Notable development in the last few years include, LeNet-5 (1998): The first successful implementation of CNN was by Yann LeCun, who developed LeNet-5, a convolutional neural network for handwritten digit recognition. This model used convolutional layers and pooling layers, and achieved state-of-the-art performance on the MNIST dataset. AlexNet (2012): In 2012, Alex Krizhevsky, Ilya Sutskever, and Geoffrey Hinton developed AlexNet, which won the ImageNet Large Scale Visual Recognition Challenge (ILSVRC) by a large margin. This model consisted of 5 convolutional layers, 3 max-pooling layers, and 3 fully-connected layers. AlexNet was notable for its use of ReLU activation functions, dropout regularization, and data augmentation techniques. VGGNet (2014): The VGGNet architecture, developed by Karen Simonyan and Andrew Zisserman in 2014, used a much deeper architecture than previous models. VGGNet had up to 19 layers, with all layers being either convolutional or max-pooling layers, followed by a few fully connected layers. This architecture achieved very high accuracy on the ImageNet dataset, and was a key step towards deeper neural networks. ResNet (2015): In 2015, Kaiming He, Xiangyu Zhang, Shaoqing Ren, and Jian Sun introduced ResNet, a CNN architecture that was even deeper than VGGNet. ResNet used residual connections to address the vanishing gradient problem that arises in very deep networks. This architecture achieved state-of-the-art performance on several computer vision benchmarks, including ImageNet and CIFAR. Inception (2014): The Inception architecture, developed by Christian Szegedy et al. in 2014, introduced the idea of using multiple filter sizes in a single convolutional layer, allowing the network to capture features at different scales. The architecture also included 1x1 convolutional layers to reduce the number of parameters in the network. The Inception architecture achieved state-of-the-art performance on the ImageNet dataset, and has since been extended in various ways. DenseNet (2016): In 2016, Gao Huang, Zhuang Liu, and Kilian Q. Weinberger introduced DenseNet, a CNN architecture in which each layer received input from all preceding layers. This created a densely connected architecture that could improve gradient flow and reduce the number of parameters. DenseNet achieved state-of-the-art performance on several image classification benchmarks, and has been used in a variety of applications.

\begin{table}[h!]
  \begin{center}
    \normalsize\addtolength{\tabcolsep}{2pt}
    \def\arraystretch{1.4}    
    \begin{tabular}{ |m{6cm}|c|c| } 
      \hline
      \textbf{Major Development}  & \textbf{Year} & \textbf{Contributed By}  \\ 
      \hline
      Visual nervous system model. & 1962,1965 & Hubel and Wiesel  \\ 
      \hline
      Mechanism of feature extraction in the visual nervous model. & 1970,1971 & K. Fukushima  \\
      \hline
      Cognitron : A self-organized multi-layer neural network. & 1975 & K. Fukushima \\
      \hline
      Neocognitron : A self-organizing neural network model for a mechanism of pattern recognition unaffected by shift in position. & 1980 & K. Fukushima  \\ 
      \hline
      Derivation of back propagation based on the Lagrangian formalism. & 1988 & Yann LeCun \\
      \hline
      Object Recognition with gradient-based learning. & 1999 & Yann LeCun   \\
      \hline
      First GPU implementation of CNN. & 2006 & K. Chellapilla et al. \\
      \hline
      Flexible and fully online implementation of CNN on GPU. & 2011 & Dan Ciresan \\
      \hline
      
    \end{tabular}
  \end{center}
  \caption{Major development in the history of CNN.}
  \label{tab:devlopmentable}   
\end{table}

 \section{Working of CNN} 
In CNN, we take an image input, and assign importance to its various features in the image and every image is uniquely identified based on those features. A CNN mainly has three layers: a convolutional layer, a pooling layer, and a fully connected layer. Padding is also an optional step used for controlling the size of the output image.

\subsection{Convolution layer}
This layer does most of the computational work. In the convolutional layer, parameters are set of learnable filters. Learnable because they change as we start processing the training data. Every filter has equal width and height with full depth of the input image. For example, 8 $\times$ 8 $\times$ 4 filters have height and width of 8 and 4 as images have CMYK color channel. As the convolution layer uses learnable filters it preserves the relationship among input pixels. This layer performs a dot product between two matrices, where one matrix is the filter/kernel and the other matrix is the portion of the image. To visualize operations of this layer we assume the input of the current layer is  $I^{input}$ that have $c^{n} \text{ channels of } a^{n} \times b^{n} \text{ size}$ 

\[
\begin{bmatrix}
  I_{1,1,1} & \dots &  I_{1,b^{n},1} \\
  \vdots   & \ddots & \vdots \\
   I_{a^{n},1,1} &    &  I_{a^{n},b^{n},1}
 \end{bmatrix}
 \dots
\begin{bmatrix}
  I_{1,1,d^{n}} & \dots &  I_{1,b^{n},d^{n}} \\
  \vdots   & \ddots & \vdots \\
   I_{a^{n},1,d^{n}} &    &  I_{a^{n},b^{n},d^{n}}
\end{bmatrix}
\]

\vspace{1cm}
The goal is to generate an feature matrix $ I^{out}$ $\text{of size } a^{out} \times b^{out} $  \text{ To generate the output, we consider a filter } f  \text{ which is a 3-D weight matrix of size } $h \times h \times d^{n} $

 This filter would have the following matrices \text{and bias term } b.

\[
\begin{bmatrix}
  f_{1,1,1} & \dots &  f_{1,h,1} \\ %
  \vdots   & \ddots & \vdots \\
   f_{h,1,1} &    &  f_{h,h,1}
 \end{bmatrix}
 \dots
\begin{bmatrix}
  f_{1,1,d^{n}} & \dots &  f_{1,h,d^{n}} \\
  \vdots   & \ddots & \vdots \\
   f_{h,1,d^{n}} &    &  f_{h,h,d^{n}}
\end{bmatrix}
\]

\vspace{1cm}

By convolutional operation, we want to extract the local information of the image. We perform the convolutional operation by doing the inner product of the sub-image and a filter. For the applying filter, we are going to obtain small regions of size the equal size of the filter that is h $\times$ h and calculates the inner product between all of the regions of the image and the filter. For example, if we start from the upper left corner of the input image first sub-image of the channel d is

\[
\begin{bmatrix}
  I_{1,1,d} & \dots &  I_{1,h,d} \\ %
  \vdots   & \ddots & \vdots \\
  I_{h,1,d} &    &  I_{h,h,d}
\end{bmatrix}
\]

Then to do a convolutional operation we solve the following.

\[
  \sum_{d=1}^{d^{n}}
  \left\langle\
  \begin{bmatrix}
  I_{1,1,d} & \dots &  I_{1,h,d} \\ %
  \vdots   & \ddots & \vdots \\
  I_{h,1,d} &    &  I_{h,h,d}
\end{bmatrix}
\cdot
\begin{bmatrix}
  f_{1,1,d} & \dots &  f_{1,h,d} \\
  \vdots   & \ddots & \vdots \\
   f_{h,1,d} &    &  f_{h,h,d}
 \end{bmatrix}
\right\rangle
+
b, \text{ where } b \text{ is bias for filter}.
\]

This value becomes the (1,1) position of the featured matrix.Now we need to get other sub-images to produce values for other positions of the output image. We specify the strides for sliding the filter. That is, we move s pixels vertically or horizontally to get sub-images. For the (2, 1) position of the output image, we move down s pixels vertically to obtain the following sub-image:

\[
\begin{bmatrix}
  I_{1+s,1,d} & \dots &  I_{1+s,h,d} \\ %
  \vdots   & \ddots & \vdots \\
  I_{h+s,1,d} &    &  I_{h+s,h,d}
\end{bmatrix}
\]

Then to get the value at the (2,1) position of the feature matrix 

\[
  \sum_{d=1}^{d^{n}}
  \left\langle\
\begin{bmatrix}
  I_{1+s,1,d} & \dots &  I_{1+s,h,d} \\ %
  \vdots   & \ddots & \vdots \\
  I_{h+s,1,d} &    &  I_{h+s,h,d}
\end{bmatrix}
\cdot
\begin{bmatrix}
  f^{n}_{1,1,d} & \dots &  f^{n}_{1,h,d} \\
  \vdots   & \ddots & \vdots \\
   f^{n}_{h,1,d} &    &  f^{n}_{h,h,d}
 \end{bmatrix}
\right\rangle
+
b, \text{ where } b \text{ is bias for filter}.
\]

In this way, we are going to get the whole feature matrix. The size of the feature matrix can be determined by the following

\[a^{out} = \left \lfloor{a^{n}-h \over s} \right \rfloor + 1, b^{out} = \left \lfloor{b^{n}-h \over s} \right \rfloor + 1 \]

An additional step while applying the filter is padding. This is done as sometimes the filter does not fit properly for a given input image. We then either pad zeros into the picture(called zero-padding ) or just drop the image data where the filter is not fitting (called valid padding).

Later on, making the decision to fire a neuron or not, less computation heavy, we use the activation function on the current feature matrix. We have multiple activation functions available. Some of them are Sigmoid, Tanh, and ReLu. Each of the activation functions has its own merits and demerits. The activation function should be decided on the bases of the data we are going to input into CNN.

For CNN, generally following ReLu activation function is used $\sigma(x) = max(x,0)$. We consider our feature matrix as $I^{f}$ till now. Then we can get $I^{out}$ matrix by $I^{out} = \sigma(I^{f})$

\subsection{Pooling}
We take the final $I^{out}$ feature matrix from the previous layer and do a pooling operation on it. This is done to reduce the computation cost later on if the image is too large. Generally, pooling operation that can extract rotational and transitional invariance features is given preference. Spatial pooling reduces the dimensionality of input but (approximately) retains important information. There are mainly three types max pooling, average pooling, sum pooling

Generally, max pooling is used in CNN. In max pooling largest element from feature, matrix is taken. After pooling, we gave a feature matrix that is of a smaller size.

\subsection{Fully connected layer}
Till now we have just extracted the features we are interested in and haven't done anything to classify images. Here fully connected layers come into the picture. We flatten our matrix into a vector and feed it into the feed-forward neural network. Then backpropagation is applied to every iteration of training.

\section{Contribution and Proposed Algorithm }

\textbf{Contributions}:
In this paper, we propose modifications to the Newton-CG method for CNN. We take whole training data instead of a very small subset of the training data. This will increase the accuracy of the next step while optimizing the function. We also consider the fact that mini-batches of training data can be processed in parallel. This can decrease our running time which would have gotten worse because of the increase in the sample size of the training data. \\
\textbf{Proposed method:}
In the purposed method instead of the sub-sampled Newton method where a subset S of the training data is used to derive the sub-sampled Gauss-Newton matrix, We make the process of deriving the Gauss-Newton matrix more accurate by taking the whole training data. This will make the task more computation hungry but we can make it better time-wise by taking advantage of the multi-cores of the processors. Taking the whole training data will increase the accuracy of steps we take to find the minima and we can converge in less time. This will in return can make our network faster. There will be a trade-off between computational power and time. We are trading computational power for a better time but this can be used in situations where we need to prioritize time over computational power like in a real-time system. In these types of systems, the function is needed to be performed in a specific time frame. A real-time application where training is done while running can take the benefit of having higher computational power for completing the work in less time. In the future, another modification we are proposing is to make the process of finding gradient distributed. This is possible as taking derivative is an independent and atomic operation. We can make batches of operations to find derivatives and run each batch on a single core in parallel with other batches. This will make the process of finding the gradient faster.

\begin{figure*}
  \begin{center}
    \includegraphics[scale=0.9]{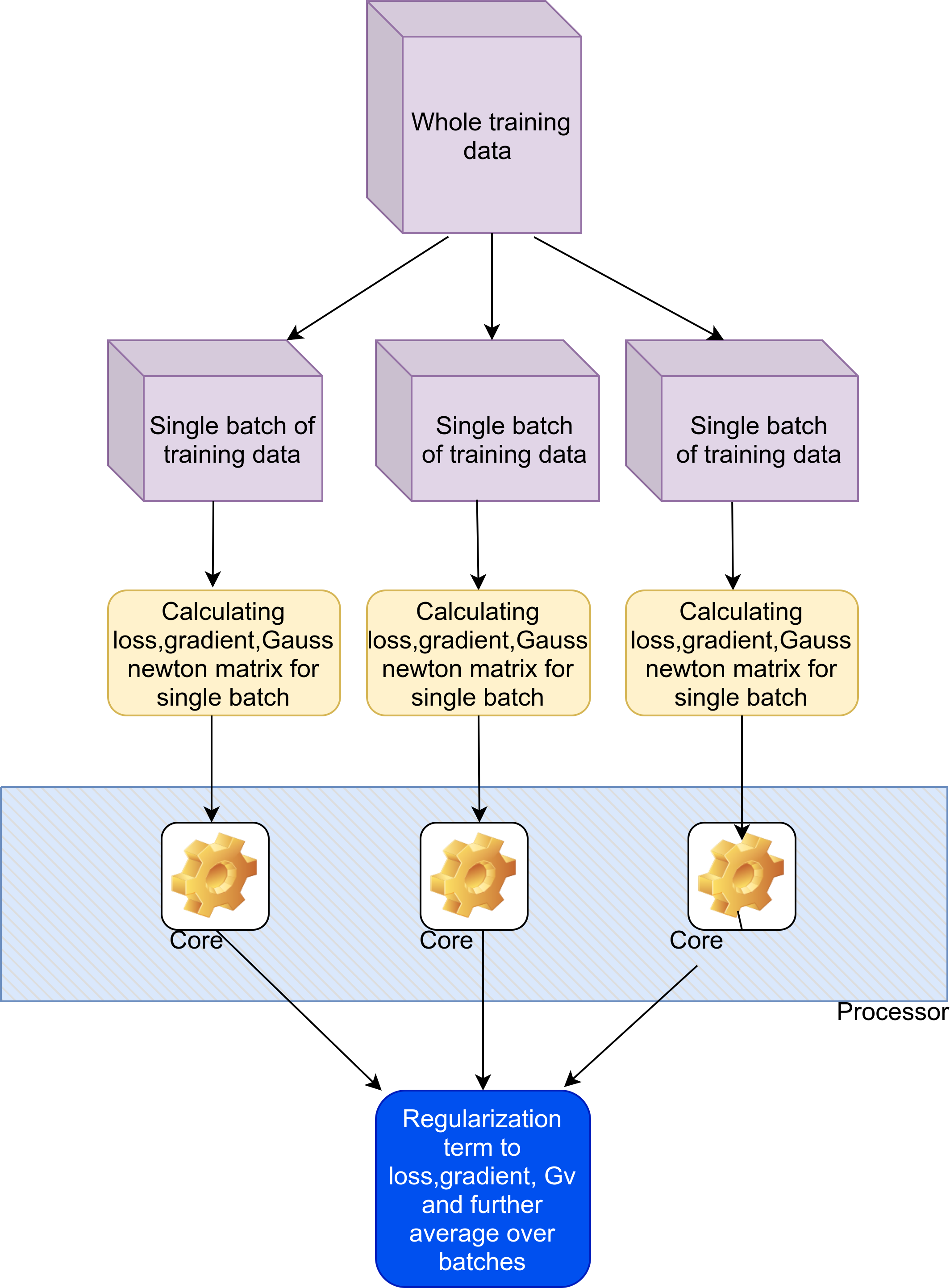} \\
  \end{center}
  \caption{How multiple cores are being used}
  \label{fig:1}
\end{figure*}

\begin{algorithm*}
  \caption{An algorithm for Hessian method in CNN}\label{Proposed Algorithm}

\begin{algorithmic}
  \State{We have $\theta$. Where $\theta$ is a collection of all the filters and Weights/biases for a fully connected layer. Calculate  $f(\theta)$.}

  \While{$\nabla f(\theta) \neq 0$}

  \State{Compute $\nabla f(\theta)$ and the needed information for Gauss-Newton matrix-vector products.Obtain a direction $d$ \newline toward function reduction. Here we can use multi-threading to compute $\nabla f(\theta)$. If we have function $f(\theta)$ as $f(x,y,...)$. We are going to compute the gradient by       
          
     $ \nabla f(\theta) =  \frac{\partial f}{\partial x} ,\frac{\partial f}{\partial y}.......$    Where $\theta$ is collection of all the filters and Weights/biases.
    
}

\State {Here we can distribute the load of finding $\frac{\partial f}{\partial x}$, $\frac{\partial f}{\partial y}$ ..... by applying parallel processing on all the partial derivatives.    }
  
  \State{ $\alpha = 1$};
 
 \While{true}
   \State{Compute $f(\theta + \alpha d)$;}

   \If{$f(\theta + \alpha)d \leq f(\theta) + \mu\alpha \nabla f(\theta)d$}
   \State{break;}
   \EndIf

   \EndWhile
   
   \State{Update $\lambda$ }
   \State{$ \theta \gets \theta + \alpha  d$}
 
 \EndWhile

\end{algorithmic}  
\end{algorithm*}

In the above algorithm, we have a variable $\theta$ that has all the variables we need to change in order to train our CNN and make it suitable for real usage. The value these variables hold is a core part of any neural network. We take all these variables into a function and minimize that function.  

As we need to solve a linear system with a large number of variables to find the next step toward minima, the Sub-sampled Hessian Newton method was used \cite{7226530}. Where data points are assumed to be from the same distribution and due to this they can be reasonably approximated by selecting a subset of whole data. We continue till we have $\nabla f(\theta) \neq 0$ which implies this is going to run till we have reached the minima \cite{deri}. After this, we would have chosen the subset of the data according to the original implementation but we will be taking the whole of the training data to have a more accurate approximation and deal with processing complexity using multiple cores to solve simultaneously. Then we compute the gradient and solve the linear system by CG to obtain a direction \textit{\textbf{d}}. We have taken $\alpha$ constant as 1. We focus on computing $f(\theta + \alpha d)$ and update $\alpha$ with $\alpha = \frac{\alpha}{2}$ until $ f(\boldsymbol{\theta}+\alpha \boldsymbol{d}) \leq f(\boldsymbol{\theta})+\eta \alpha \nabla f(\boldsymbol{\theta})^{T} \boldsymbol{d} $ is satisfied . Then We update $\lambda$ based on how good the function reduction is $\rho$ is the ratio between the actual function reduction and the predicted reduction. Using $\rho$, the parameter $\lambda_{next}$ for next iteration is decided by

$$ \lambda_{\text {next }}=\left\{\begin{array}{ll}
\lambda \times \text { drop }  \rho>\rho_{\text {upper }} \\
\lambda \ \   \rho_{\text {lower }} \leq \rho \leq \rho_{\text {upper }} \\
\lambda \times \text { boost }  \text { otherwise }
\end{array}\right.
$$

Then we update our $\theta$ with $$
\theta \leftarrow \theta+\alpha d
$$

\section{Experiment and Results}
We have primarily tested on data set MNIST\cite{mnist} and CIFAR10\cite{cifar}. MNIST is a database of handwritten digits with a training set of 60,000 Grey scale images of size 28 $\times$ 28, and a test set of 10,000 same-sized images and CIFAR-10\cite{cifar} dataset consists of 60000 32x32 color images in 10 classes, with 6000 images per class. There are 50000 training images and 10000 test images. When we run the original implementation to find the time taken per Gv iteration as we increase the size of the sample for approximating the Gauss Newton Matrix. We then plot these values on the graph. Where Y-axis represents the size of the sample taken for the approximation and the X-axis shows the time taken for the one iteration.

\begin{center}

\captionof{}{Running on MNIST}

\begin{tabular}{||c c c||} 
  \hline
  Data Size & Without Threads & With Threads \\ [0.5ex] 
  \hline\hline
  64000 & 189.14  & 9.85 \\ 
  \hline
  32000 & 104.49 & 5.20 \\
  \hline
  1600 & 53.34 & 3.16 \\
  \hline
  8000 & 25.93 & 1.74 \\
  \hline
  4000 & 14.06 & 1.56 \\
  \hline
  2000 & 7.11 & 1.43 \\ [1ex] 
  \hline

\end{tabular}
\end{center}

\begin{figure*}
  \begin{center}
    \includegraphics[scale=0.7]{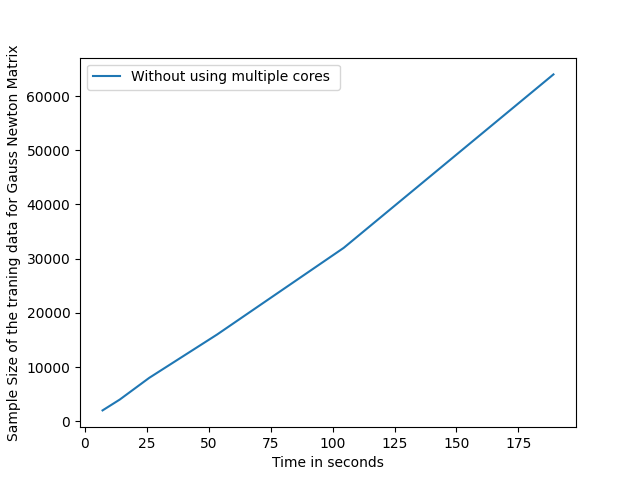}
    \end{center}
  \caption{Running on MNIST}
  \label{fig:2}
\end{figure*}

In the figure \ref{fig:2} we have plotted the time taken as we increase the sample size when working with the MNIST dataset\cite{mnist}. We can clearly observe that it is a linear line and the time increase is proportional to the increase in the sample size.

\begin{figure*}
  \begin{center}
    \includegraphics[scale=0.7]{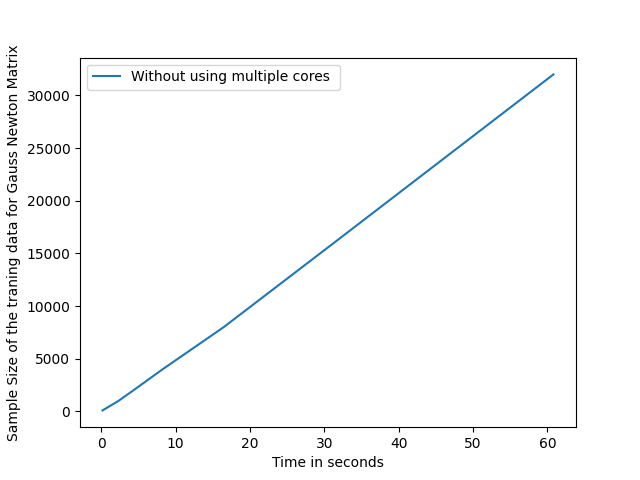}
    \end{center}
  \caption{Running on CIFAR-10}
  \label{fig:3}
\end{figure*}

In the figure \ref{fig:3} we have plotted the time taken as we increase the sample size when working with the CIFAR-10\cite{cifar} dataset. We clearly observe that it's also a linear line and the increase in time taken is directly proportional to the increase in the sample size.

We got this data by running the original implementation code shared by the author. We ran the implementation with changing data size for Gauss-Newton Approximation and with different databases. To make understanding clear let us have look at the command line arguments given.


\begin{center}

\captionof{}{Running on CIFAR}

\begin{tabular}{||c c c||} 
  \hline
  Data Size & Without Threads & With Threads \\ [0.5ex] 
  \hline\hline
  64000 & 60.81133  & 28.4656 \\ 
  \hline
  32000 & 31.28563 & 27.96621 \\
  \hline
  1600 & 16.46787 & 25.32876 \\
  \hline
  8000 & 8.25654 & 23.77603 \\
  \hline
  4000 & 4.32667 & 23.55 \\
  \hline
  2000 & 2.32718 & 20.9896 \\ [1ex] 
  \hline

\end{tabular}
\end{center}

\begin{figure*}
  \begin{center}
    \includegraphics[scale=0.7]{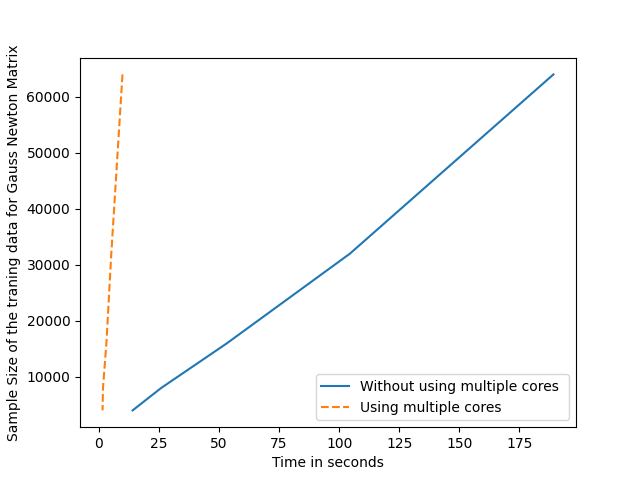}
    \end{center}
  \caption{Comparison of old and new implementation on MNIST}
  \label{fig:6}
\end{figure*}

\begin{figure*}
  \begin{center}
    \includegraphics[scale=0.7]{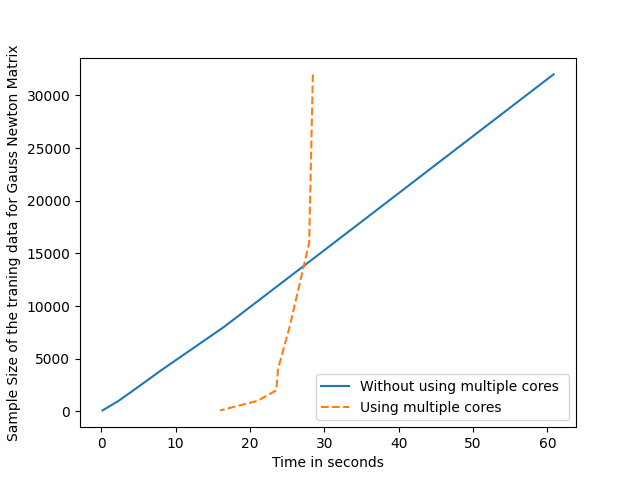}
    \end{center}
    \caption{Comparison of old and new implementation on CIFAR-10}
  \label{fig:7}
\end{figure*}

We have run the python program for training. Then this program takes a few arguments first one is \textit{--optim} it's there to choose the optimizer as we are working specifically using NewtonCG we are not going to make any modifications to it. The next argument is where we are interested in \textit{--GNsize} defines how many samples of the training data are to be used in the approximation of the Gauss-Newton matrix. The next argument we are interested in is \textit{--train\_set} where we give a path to the file containing training data. Then comes \textit{--dim} argument where we give the dimensions of the data. If we have training data containing an image of size 28 $\times$ 28 then dimensions are going to be 28 28 1. We can think of the last one as the number of bytes required to represent each pixel. A colored image with the same dimensions will be given \textit{--dim} as 28 28 3. As it takes 3 bytes to represent a single pixel in the RGB image. Now when we run these commands we get results that are plotted.
We got our data from this log output of the program. First with the original implementation and then with our changes. When we plotted  the data we get figure \ref{fig:6} and \ref{fig:7}. As we can see in the figure \ref{fig:6} we have got a better time from the start in comparison to the original implementation. While on CIFAR-10\cite{cifar} original implementation is faster at the start but gets slower as the sample size increase.

\section{Conclusion and Future work}
We can conclude that using parallel processing for computations can decrease our processing time without making drastic changes to the older implementation. When we specifically instructed computations to be done on multiple cores simultaneously it took less time to complete the task. As those operations are independent of each other it wouldn't affect the result. We can apply this parallel processing in all the tasks where we do not take any external input during the calculation, which can change dynamically. In the future, we can go one step further and similarly calculate the gradient with each partial derivative being an independent atomic operation. This can increase CPU and GPU utilization resulting in less time to complete those tasks.

\bibliography{references}{}
\bibliographystyle{plainurl}

\end{document}